\documentclass[runningheads]{llncs}
\usepackage{graphicx}
\usepackage{multirow} 
\usepackage{setspace}
\usepackage{microtype}

\usepackage{xcolor}

\begin{document}
\title{Predicting Themes within Complex Unstructured Texts: \\ A Case Study on Safeguarding Reports}
\titlerunning{Predicting Themes within Complex Unstructured Texts}
%
\author{Aleksandra Edwards\inst{1,2} \and
David Rogers\inst{1,2}\and
Jose Camacho-Collados\inst{1}\and
H\'el\`ene de Ribaupierre\inst{1}\and
Alun Preece\inst{1,2}}
\authorrunning{A. Edwards et al.}
%
\institute{School of Computer Science and Informatics, Cardiff University, Cardiff, UK \\\and
    Crime and Security Research Institute,Cardiff University, Cardiff, UK \\}
\maketitle              
\begin{spacing}{0.90}
\begin{abstract}
Text classification typically requires large amounts of labelled training data; however, the acquisition of high volumes of labelled datasets is often expensive or unfeasible, especially for highly-specialised domains for which both training data (documents) and access to subject-matter expertise (for labelling) is limited. Language models pre-trained on large amounts of text corpora provide state-of-the-art performance against most standard natural language processing (NLP) benchmarks, including text classification. However, their prevalence over more traditional linear classifiers and domain-based approaches have not been investigated fully.
In this paper, we address the combination of state-of-the-art deep learning and classification methods and provide an insight into what combination of methods fit the needs of small, domain-specific, and terminologically-rich corpora. We focus on a real-world scenario related to a collection of safeguarding reports comprising learning experiences and reflections on tackling serious incidents involving children and vulnerable adults. Our aim is to automatically identify the main themes in a safeguarding report using three main types of classification. Our results show that for a very small amount of data a simple linear classifier outperforms state-of-the-art language models. Further, we show that the performance of classifiers is more affected by the size of the training data rather than the amount of context given.

\end{abstract}

\section{Introduction}\label{sec:intro}
\vspace{-0.2cm}
The performance of natural language processing (NLP) classification tasks is heavily reliant on the amount of training data available \cite{turker2019knowledge}. However, the acquisition of high volumes of labelled data can be an expensive, time- and resource-consuming process \cite{ali2019text}, especially when the text to be labelled is in a highly-specialised domain where only scarce domain experts can perform the manual labelling task \cite{turker2019knowledge}. Current pre-trained neural models such as BERT (Bidirectional Encoder Representations from Transformers) \cite{Devlin2019BERTPO} proved to provide state-of-the-art results in most standard NLP benchmarks \cite{wang2018glue}, including text classification. However, the applicability of these language models to very small collections of highly specialised documents has not been fully explored or compared to more traditional methods. A limitation to pre-trained models is that there is still a need for task-specific datasets for these models to perform well in a specific domain \cite{radford2019language}. Therefore, adapting these large but generic models to specific domains and tasks has become the new standard approach for many NLP problems~\cite{sainz2021ask2transformers}. For instance, the authors of~\cite{gururangan2020don} provide a more extensive research on  whether it is still helpful to tailor a pretrained model to the domain of a target task. However, this research is not focused on text classification and does not compare neural models to other types of machine learning models. Further,
a recent research~\cite{edwards2020go} on few-shot classification, analysed the role of labeled and unlabeled data for classifiers by comparing a linear model such as fastText coupled with domain-specific embeddings against fine-tuned BERT model using both domain-specific and generic corpora. However, the authors performed analysis using generic datasets assuming the presence of large amounts of unlabeled data which can be used for fine-tuning models on domain data. We build on this research \cite{edwards2020go} by
comparing the performance of three types of classifiers for a real-world scenario related to the safeguarding domain where there is very limited amount of labeled and unlabeled dataset. Previous work on performing NLP analysis on the safeguarding corpus emphasized the challenges of extracting knowledge from the documents using off-the-shelf text analysis tools due to the highly specialised lexical characteristics of the reports~\cite{edwards2019knowledge}. Further, there is no existing knowledge resources which fit the needs of the domain~\cite{edwards2019knowledge} which makes the use of semantic enrichment approaches for the domain difficult.

We look at whether domain-trained embeddings are effective even when trained on very limited corpus. Our main contribution is that we conduct a thorough analysis of what combination of embedding and language models and classification approaches fit the needs of a small domain-specific and terminology-rich corpus. Further, we also look at how deep learning approaches are affected by training dataset size versus the amount of context given. 


\section{Case study: Safeguarding reports}\label{sec:case}
\vspace{-0.2cm}
The purpose of a safeguarding report is to identify and describe related events that precede a serious safeguarding incident --- for example, involving a child or vulnerable adult --- and to reflect on agencies' roles and the application of best practices. Each report contains key information about learning experiences and reflections on tackling serious incidents. The reports carry great potential to improve multi-agency work and help develop better safeguarding practices and strategies \cite{edwards2019knowledge}. Analyzing and understanding safeguarding reports is crucial for health and social care agencies; in particular, a key task is to identify common themes across a set of reports. Traditionally, this is done in social science by a process of manually annotating the reports with themes identified by subject-matter experts using a qualitative analysis tool such as NVivo.
However, each report is lengthy and complex, so manual annotation is a time-consuming and potentially bias-prone process \cite{edwards2019knowledge}. Furthermore, in our particular case, the safeguarding collection is expected to grow significantly in the near future, with the additional resourcing of 500 historical reports, making the manual annotation of these additional documents unfeasible. Therefore, we aim to automate the process of document annotation.

Furthermore, in our particular case, the safeguarding collection is expected to grow significantly in the near future, with the additional resourcing of 500 historical reports, making the manual annotation of these additional documents unfeasible. Therefore, we aim to automate the process of document annotation. 

The thematic framework~\cite{robinson2019making} used for performing document classification resulted from collaborative work between multiple subject-matter experts.  In this context, a \emph{theme} refers to the main topic of discussion related to safeguarding incidents, specifically relevant to domestic homicide and mental health homicide.

\section{The Dataset}\label{sec:data}
\vspace{-0.2cm}

    At the time of development the corpus consisted of 27 full safeguarding reports. The annotations were carried out by a social science team following standard methodology in the field. They used a qualitative analysis tool (NVivo) to label parts of documents with thematic annotations from 5 top-level themes according to the thematic framework described in Section \ref{sec:case}. The annotation was performed by labelling different-length passages of the reports with themes from the thematic framework. The majority of report contents were labeled except appendices.
    The total number of sentences in the corpus was 3,421 (see Table~\ref{tab1}) with unbalanced distribution between the different themes where sentences can be associated with multiple themes. 
          \begin{table}[hbt!]
           \caption{Data distribution of sentences per themes}\label{tab1}
		    \centering
		        \large
		      \resizebox{\linewidth}{!}{
		       \begin{tabular}{|p{3.0cm}|c|c|c|p{7.5cm}|p{7.5cm}|}\hline
		        \textbf{Theme}&\textbf{Train}&\textbf{Dev}&\textbf{Test}&\textbf{Description}&\textbf{Example}\\\hline\hline
		        Contact with Agencies&1,281&335&219&Agency interactions with the people involved prior to the incident&The person injured his ankle and was seen at the GP surgery\\\hline
		        Indicative \newline Behaviour&1,078&276&83&Types of behaviour that might indicate a risk to self and others, such as signs of aggression, previous offences&The perpetrator had a long history of alcohol misuse and criminality\\\hline
		        Indicative\newline Circumstances&427&104&99&Personal circumstances prior the incident that might indicate a risk to self and others, such as relationship problems, debt&Their relationship was based on the economic realities of subletting a flat informally\\\hline
		        Mental Health Issues&316&76&51&Indications of any mental health problems that anyone involved in the incident experienced&They were diagnosed with Attention Deficit Hyperactivity Disorder\\\hline
		       Reflections&780&203&78&Key lessons learned in reviewing the case&This highlights a challenge for agencies on what information to share when victims and perpetrators reside in different administrative areas\\\hline
		        Total&2,736&685&300&&\\\hline
		        \end{tabular}
		       }
		      
		   \end{table}
    We evaluated models performance using training, development (both training and development were randomly sampled from the 27 reports)  and test sets. 
    Both development and test sets were annotated at the passage-level. 
    The test set was extracted from safeguarding reports different from the original 27 documents. The test set contained a 100 randomly selected passages where each passage consisted of 3 sentences. Due to the limited amount of reports available, we built and evaluated classifier models on a sentence level (i.e., results presented in Section \ref{sec:res}). 
    Thus, each sentence was assigned the label of the passage to which it belongs. Further, we ensured that the train and development set do not intersect by automatically selecting random non-overlapping partitions for the two subsets. We also performed analysis at the passage-level, presented in Section \ref{sec:erroranalysis}.

\section{Classification Experiments}\label{clexepirments}
\subsection{Methods}\label{sec:methods}
\vspace{-0.2cm}

In our experiments, we perform multi-label classification to identify main themes within documents. We compare three classifiers --- a simple count-based classifier, a linear classifier based on word embeddings, and a state-of-the-art language model. We perform experiments with pre-trained and corpus-trained embeddings as well as different methods for building feature vectors. We use an n-gram feature representation and a Naive Bayes classifier as our baseline. 
    
Our method consists of four overall steps, described below. 
     \begin{figure}[hbt!]
     \centering
	    \includegraphics[scale = 0.09]{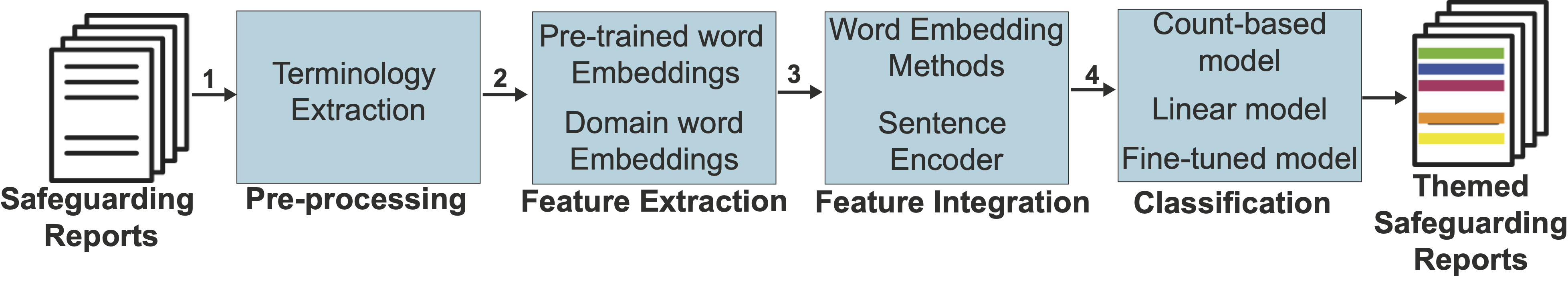}
	  
        \caption{Method overview}\label{fig1}
    \end{figure}
    \paragraph{Step 1: Pre-processing}\label{ssec:stepone}
    We extracted terms from the corpus using FlexiTerm \cite{spasic2013flexiterm}, an open-source software tool for automatic recognition of multi-word terms. We used the sentences pre-processed with the terminology extraction step for building sentence embeddings and for creating simple n-gram feature vectors. 
        
    \paragraph{Step 2: Feature Extraction (FE)}\label{ssec:steptwo}
    We used fastText word embeddings \cite{bojanowski2017enriching} pre-trained with subword information on Common Crawl.  We also used fastText for learning domain-specific embeddings because it captures the meaning of rare words better than other approaches. We used the skip-gram method for building word embeddings with 300 dimensions.

    \paragraph{Step 3: Feature Integration (FI)}\label{ssec:stepthree}
        We use several ways for combining the word embeddings into reduced sentence representations: In the first approach, we average the embeddings of each word in a sentence along each dimension.  In the second approach, we assign TF-IDF weights to the words in a sentence, and calculate the weighted average of the word embeddings along each dimension (where the contribution of a word is proportional to its TF-IDF weight).
            
         Finally, we use Bidirectional Encoder Representations from Transformers (BERT) \cite{Devlin2019BERTPO}. A limitation of the word embedding model described above is that it produces a single vector of a word despite the context in which it appears. In contrast to the other embedding methods, BERT is designed to pre-train deep bidirectional representations from unlabeled text by jointly conditioning on both left and right context in all layers. This characteristic allows the model to learn the context of a word based on all of its surroundings, thus it generates more contextually-aware word representations. There are two steps in the BERT framework: pre-training and fine-tuning. In this step of the methodology, we use the base pre-trained BERT model, trained on the Books Corpus and English Wikipedia, for extracting contextualized sentence embeddings. The fine-tuning step consists of further training on the downstream tasks.
         
    \paragraph{Step 4: Classifiers}\label{ssec:stepfour}
     We perform classification on a sentence level where each sentence had been assigned the theme of the passage the sentence belonged to.  Here, we take `ground truth' to be the annotations made by the  social scientist expert annotators who were involved in creating the thematic framework (see Section\ref{sec:case}). For a \emph{baseline} we use  GNB classifier
     based on frequency-based features available in Scikit-Learn library \cite{pedregosa2011scikit}, since it is considered a strong baseline for many text classification tasks\cite{fan2008liblinear}.  A potential problem with linear classifiers is that they struggle with OOV words, fine-grained distinctions and unbalanced datasets. 
     The \emph{fastText classifier}\cite{joulin2017bag} addresses this problem by integrating a linear model with a rank constraint, allowing sharing parameters among features and classes. Further, we fine-tune BERT  for the classification task using a sequence classifier, a learning rate of 5e-5 and 4 epochs. In particular, we made use of the BERT’s Hugging Face default transformers implementation for classifying sentences \cite{Wolf2019HuggingFacesTS}.

\subsection{Results}\label{sec:res}
\vspace{-0.25cm}
    We evaluate the performance of the machine learning algorithms by using precision, recall, and F1-measure metrics. The summary results are calculated using micro- and macro- based measures.
    Early experiments using Word2Vec embeddings~\cite{mikolov2013distributed} and SVM classifier showed unsatisfactory performance compared to fastText embeddings and GNB classifier. Thus, these results are omitted from Table~\ref{tab2}.
  
    The results in Table~\ref{tab2} show that a simple terminology-based pre-processing step leads to slight improvements over the baseline with micro F1 of 0.59 in comparison to baseline micro-F1 of 0.57. Despite the small amount of data, we found that corpus trained embedding provide a notable advantage over pre-trained embeddings in the classifiers performance.
    \begin{table}[hbt!]
    	\caption{Summary classification results}\label{tab2}
        \centering
		   \scalebox{0.6}{
		   \large
		   \begin{tabular}{|p{1.9cm}|p{2.5cm}|p{2.4cm}||c|c|c|c|c|c|}\hline
		    \multicolumn{3}{|p{2.5cm}||}{\textbf{Method}}&\multicolumn{3}{c|}{\textbf{Micro}}&\multicolumn{3}{c|}{\textbf{Macro}}\\\hline
		    \textbf{Classifier}&\textbf{FE}&\textbf{FI}&\textbf{p}&\textbf{r}&\textbf{F1}&\textbf{p}&\textbf{r}&\textbf{F1}\\\hline\hline
		    Baseline&1,2 grams&count&.51&.66&.57&.48&.65&.54\\\hline\hline
		    \multirow{5}{*}{GNB}
		    &Terms&count&.52&.68&.59&.49&.66&.54\\\cline{2-9}
		    &\multirow{2}{*}{FT}&mean&.38&.54&.44&.38&.55&.42\\
		    &&TF-IDF&.27&.48&.34&.32&.56&.34\\\cline{2-9}
		    &Domain&mean&.47&.60&.53&.45&.61&.50\\\cline{2-9}
		    &BERT&BERT&.43&.60&.50&.40&.59&.47\\\hline\hline
		    \multirow{2}{*}{FT}&Domain&Mean&.52&.67&.59&.48&.62&.54\\\cline{2-9}
		    &FT&Mean&.52&.64&.57&.48&.59&.52\\\hline\hline
		    Fine-tune &BERT&BERT&\textbf{.56}&\textbf{.73}&\textbf{.64}&\textbf{.52}&\textbf{.68}&\textbf{.59}\\\hline
		  
	\end{tabular}
	}
    \end{table}
   
    fastText classifier outperformed GNB model, especially when domain-based embeddings were used.  A  non-verbatim example of a sentence where fastText model, based on corpus-trained embeddings performs better than pre-trained embedding models is: \textit{'The police received information that the subject was selling crack'}. A potential reason for fastText to classify correctly this sentence versus the classifiers using pre-trained embeddings is that the word \textit{`crack'} has the meaning of a \textit{`drug'} in the reports. However, this is not the widely accepted meaning for this word and thus it cannot be interpreted correctly by pre-trained models. The GNB based on pre-trained BERT model outperforms the classifiers based on pre-trained embeddings, however it does not lead to improvements over the domain-based models. Fine-tuning BERT is the best performing classifier with micro-F1 of 0.64 and macro-F1 of 0.59 which gives 0.5 improvement over the baseline.  
    The improvement in the results achieved by fine-tuning BERT indicate the importance of adapting even the more context-aware pre-trained language models to the specific domain, especially when the domain contains highly specialised language. Further, the poor performance of classifiers based on pre-trained word models shows the lack of transferability of pre-trained embeddings for a highly specialised domain such as the safeguarding reports.
    
    The three best-performing classifiers give similar average results between the dev and test set (see Table~\ref{tab3}). Further, models tend to return higher results for some themes, especially `Mental Health Issues' for the test set rather than the dev set. A potential reason for this may be attributed to the fact that the test set has been annotated in a similar manner to the classification models, i.e., independent of the context of the entire documents. The BERT classifier returned results above 0.50 for the themes `Contact with Agencies', `Reflections' and `Indicative Behaviour' for the dev and test datasets with precision above 0.60 and recall above 0.70. 
        
        \begin{table}[hbt!]
        \caption{Results per theme for best performing classifiers}\label{tab3}
		   \centering
		  
		    \scalebox{0.4}{
		    \huge
		    \begin{tabular}{|c|c||c|c|c|c|c|c|}\hline
		     \multirow{2}{*}{\textbf{Method}}&\multirow{2}{*}{\textbf{Theme}}&\multicolumn{3}{c|}{\textbf{dev set}}&\multicolumn{3}{c|}{\textbf{test set}}\\\cline{3-8}
		     &&\textbf{p}&\textbf{r}&\textbf{F1}&\textbf{p}&\textbf{r}&\textbf{F1}\\\hline\hline\cline{3-8}
		      \multirow{5}{*}{baseline}&Contact with Agencies&.65&.70&.68&.86&.47&.61\\
		        &Indicative behaviour&.56&.63&.59&.46&.57&.51\\
		        &Indicative circumstances&.33&.64&.44&.52&.51&.51\\
		        &Mental Health Issues&.26&.57&.36&.39&.45&.42\\ 
		        &Reflections&.58&.69&.63&.47&.76&.58\\\hline
		       &\textbf{AVERAGE}&\textbf{.48}&\textbf{.65}&\textbf{.54}&\textbf{.54}&\textbf{.55}&\textbf{.52} \\\hline\hline
		      
		         \multirow{5}{*}{FT}&Contact with Agencies&.55&.75&.63&.79&.74&.76\\
		        &Indicative behaviour&.58&.73&.65&.45&.70&.54\\
		        &Indicative circumstances&.41&.56&.47&.49&.36&.42\\
		        &Mental Health Issues&.26&.42&.33&.35&.31&.33\\ 
		        &Reflections&.58&.64&.61&.51&.64&.56\\\hline
		        &\textbf{AVERAGE}&\textbf{.48}&\textbf{.62}&\textbf{.54}&\textbf{.52}&\textbf{.55}&\textbf{.53}\\\hline\hline
		         \multirow{5}{*}{BERT}&Contact with Agencies&.62&.82&.71&.84&.58&.69\\
		        &Indicative behaviour&.60&.74&.66&.48&.63&.54\\
		        &Indicative circumstances&.47&.56&.51&.68&.34&.46\\
		        &Mental Health Issues&.31&.51&.39&.47&.46&.46\\ 
		        &Reflections&.59&.76&.67&.51&.82&.63\\\hline
		        &\textbf{AVERAGE}&\textbf{.52}&\textbf{.68}&\textbf{.59}&\textbf{.60}&\textbf{.57}&\textbf{.58}\\\hline
		     \end{tabular}
		     }
		     
		   \end{table}

\section{Error Analysis}\label{sec:erroranalysis}
\vspace{-0.3cm}
    In the preceding section we evaluated the performance of the classification approaches against the annotations generated by the creators of the thematic framework, who we refer to as the \emph{expert annotators}. By creating a classifier that uses the annotations generated by expert annotators as a `ground truth', we aim to produce unified and comparable results across generations that are not susceptible to variations in annotations created by different human annotators interpreting the coding framework. Going further, we judge the predictive power of the models by comparing their performance against the annotations of \emph{expert validators}: independent social scientists who did not participate in the creation of the thematic annotation framework.  We aim to measure the ability of the learned models to conserve the knowledge of the \emph{expert annotators} versus if the task was performed manually by independent social scientists who were not creators of the framework (Section \ref{ssec:machinevshuman}). In this way, we will be able to judge whether automated approaches are reliable for labeling the reports.   

    We perform three main types of analysis. First, we compare the performance of the classifiers against the annotations of \emph{expert validators}. 
     Secondly, we compare the performance of the classifiers for different length of sentences to observe the classifiers suitability for various sequence lengths. We also measure the effect of the training dataset size on the performance of the models (Section \ref{ssec:effect}). Thirdly and finally,  we look at the effect of the number of training instances versus the amount of context provided per instance on the performance of the classifiers (Section \ref{ssec:context}). 
    
    \subsection{Expert Validators vs Classifiers}\label{ssec:machinevshuman}
        
        The initial thematic framework was developed by annotating passages of the documents rather than individual sentences. However, our classifiers are trained with sentences. In order to fairly judge the predictive power of the models against human annotators for annotating sentences and passages of the reports, we performed a study comparing the performance of the classification models versus two independent \emph{expert validators} on sentence- and passage-level. For these purposes we used two datasets --- one consisting of sentences and one consisting of passages. The \emph{sentence set} consisted of a sample of 100 randomly chosen sentences, while the \emph{passage set} consisted of a 100 passages, each containing three sentences. The  \emph{sentence set} was extracted from the dev set while the \emph{passage set} was extracted from the test set (see Table~\ref{tab1}).  We measured the inter-annotator agreement for predicting themes using Cohen's kappa (see Table~\ref{tab4}). We also compare the average F1 measure per theme between the expert validators and the best performing classifier (BERT).

	\begin{table}[hbt!]
          \caption{Expert validator results (Cohen's Kappa, average expert F1, BERT F1): `Expert F1' refers to the average F1 measure between the two expert validators.}\label{tab4}
		   \centering
		   
		    \scalebox{0.3}{
		    \Huge
		    \begin{tabular}{|c|c|c|c|c|c|c|}\hline
		    \multirow{2}{*}{\textbf{Theme}}&\multicolumn{3}{|c|}{\textbf{sentences}}&\multicolumn{3}{|c|}{\textbf{passages}}\\\cline{2-7}\cline{2-7}
		     &\textbf{Kappa}&\textbf{Expert F1}&\textbf{BERT F1}&\textbf{Kappa}&\textbf{Expert F1}&\textbf{BERT F1}\\\hline\hline
		  
		      Contact with Agencies&.48&.56&.71&.31&.71&.72\\\hline
		        Indicative behaviour&.36&.51&.66&.16&.56&.61\\\hline
		        Indicative circumstances&.32&.39&.48&.38&.54&.58\\\hline
		        Mental Health Issues&.56&.42&.47&.67&.65&.56\\ \hline
		        Reflections&.27&.37&.65&.47&.52&.54\\\hline\hline
		       \textbf{AVERAGE}&\textbf{.40}&\textbf{.45}&\textbf{.61}&\textbf{.40}&\textbf{.60}&\textbf{.60}\\\hline\hline
		     \end{tabular}
		     }
		    
		   \end{table}	   
          The Cohen's kappa scores showed moderate agreement between the validators with an average score 0.40 on sentence and a passage level. The highest level of agreement is for `Mental Health Issues' theme. However, the average expert F1 for this theme is surprisingly low. The reason for the discrepancy between the Cohen's kappa score and the F1 measure is the occurrence of sentences which mention mental health problems such as `depression'.  Such sentences are labeled by the expert validators as `Mental Health Issues', however their true label is different because of the surrounding context. Surprisingly, a large portion of these sentences were correctly classified by BERT. The average F1 score for the expert validators significantly improves for passage-level classification with average F1 = 0.60 in comparison to sentence-level annotations where an average F1 = 0.45 (see Table~\ref{tab4}). This suggests that humans need more context --- i.e., to see the sentences embedded in paragraphs --- to classify sentences correctly, compared to deep learning models that can generalize better in these cases with limited context thanks to what they learned from the training set. 
          
    \subsection{Effect of sentence length and training size}\label{ssec:effect}
\begin{figure}[hbt!]
		    \begin{center}
		     \includegraphics[scale = 0.07]{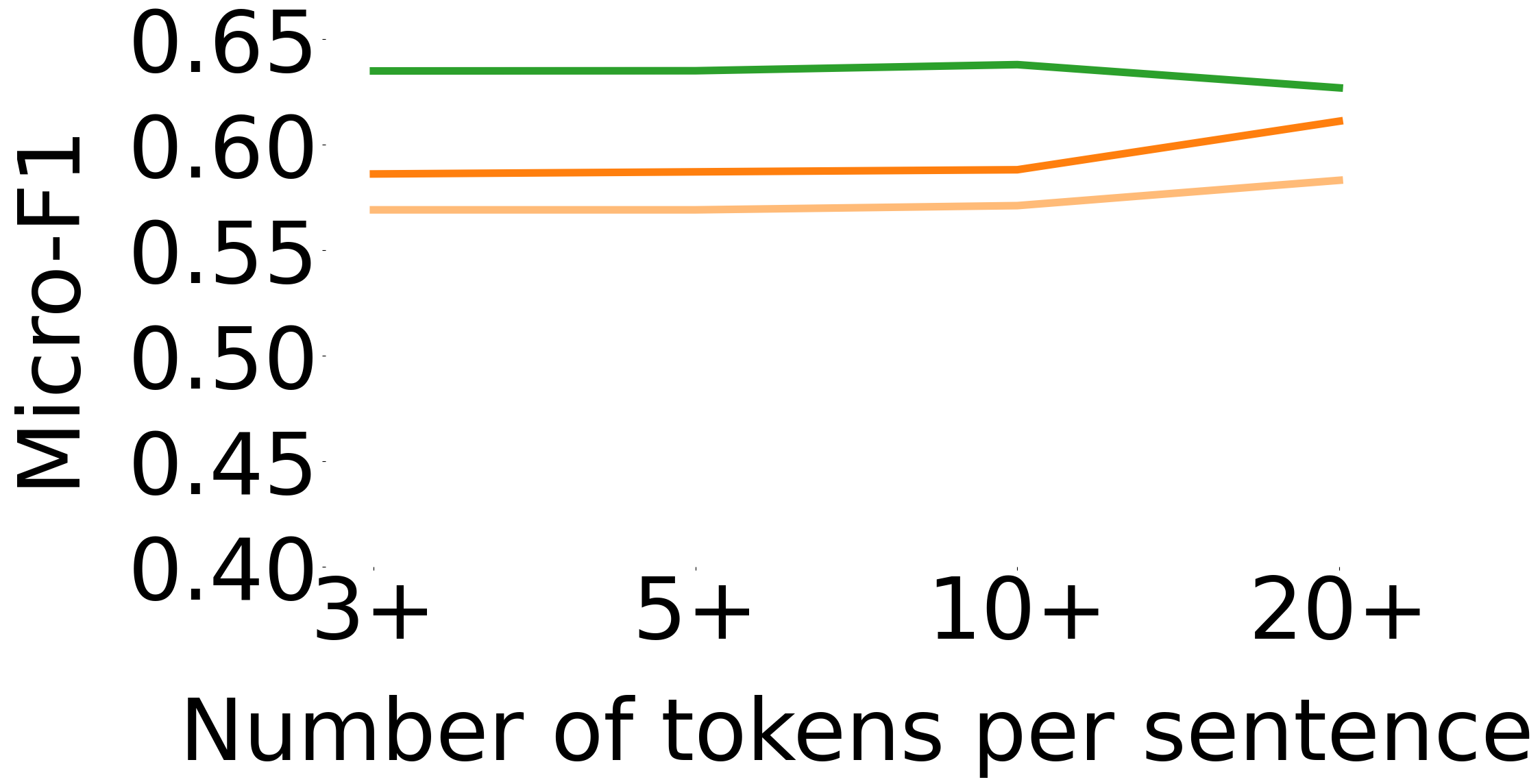}
		     \includegraphics[scale = 0.07]{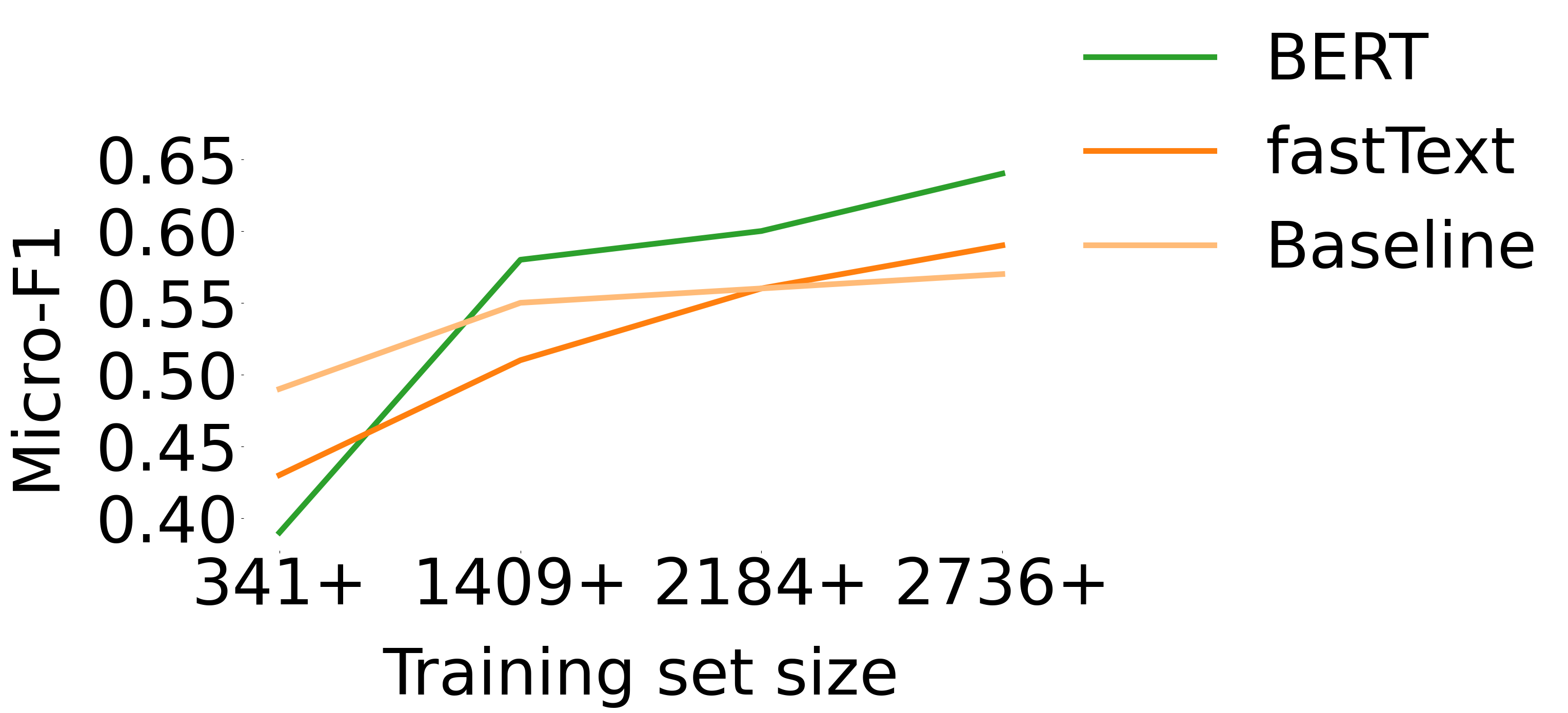}

            \caption{Micro-F1 measure per: sentence length (i.e., sent with more than 3 tokens, etc) (left) and different train dataset size (i.e., train dataset with up to 341 sentences, etc)  (right)}\label{fig3}
		    \end{center}
	       
        \end{figure}

    Experiments comparing the best-performing classifiers for different sentence length and training set size showed that BERT performed better than the baseline method for any length of sentences. Further, BERT gave higher results than fastText and the baseline for shorter sentences. For long sentences, BERT and fastText had very similar performance with a difference less than 1\% (see Fig.\ref{fig3}). The comparison between the classification models performance for different sizes of training set revealed that deep learning models (i.e., BERT) are highly influenced by the size of the training set in comparison to linear models such as the baseline and fastText (see Fig.\ref{fig3}). BERT performed worse than the baseline for the very small training set while fastText gave similar performance to the baseline. However, BERT's performance almost doubled as more sentences were added to the training set while GNB performance was not that heavily influenced by the size of the training data, especially for a training set with more than 1,000 sentences.

    \subsection{Sentences vs Passages}\label{ssec:context}
\begin{figure}[hbt!]
		\begin{center}
		     \includegraphics[scale = 0.07]{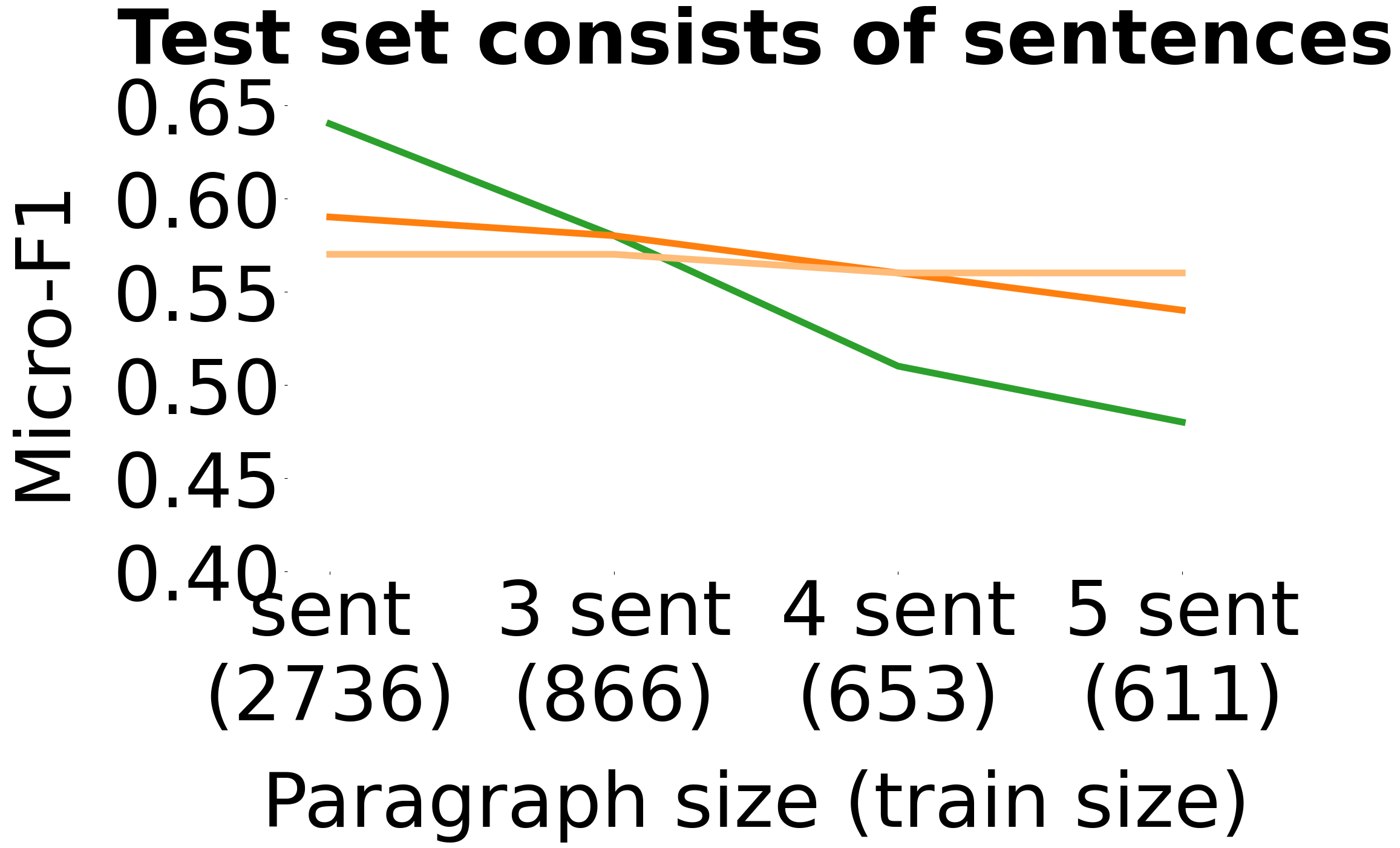}
		     \includegraphics[scale = 0.07]{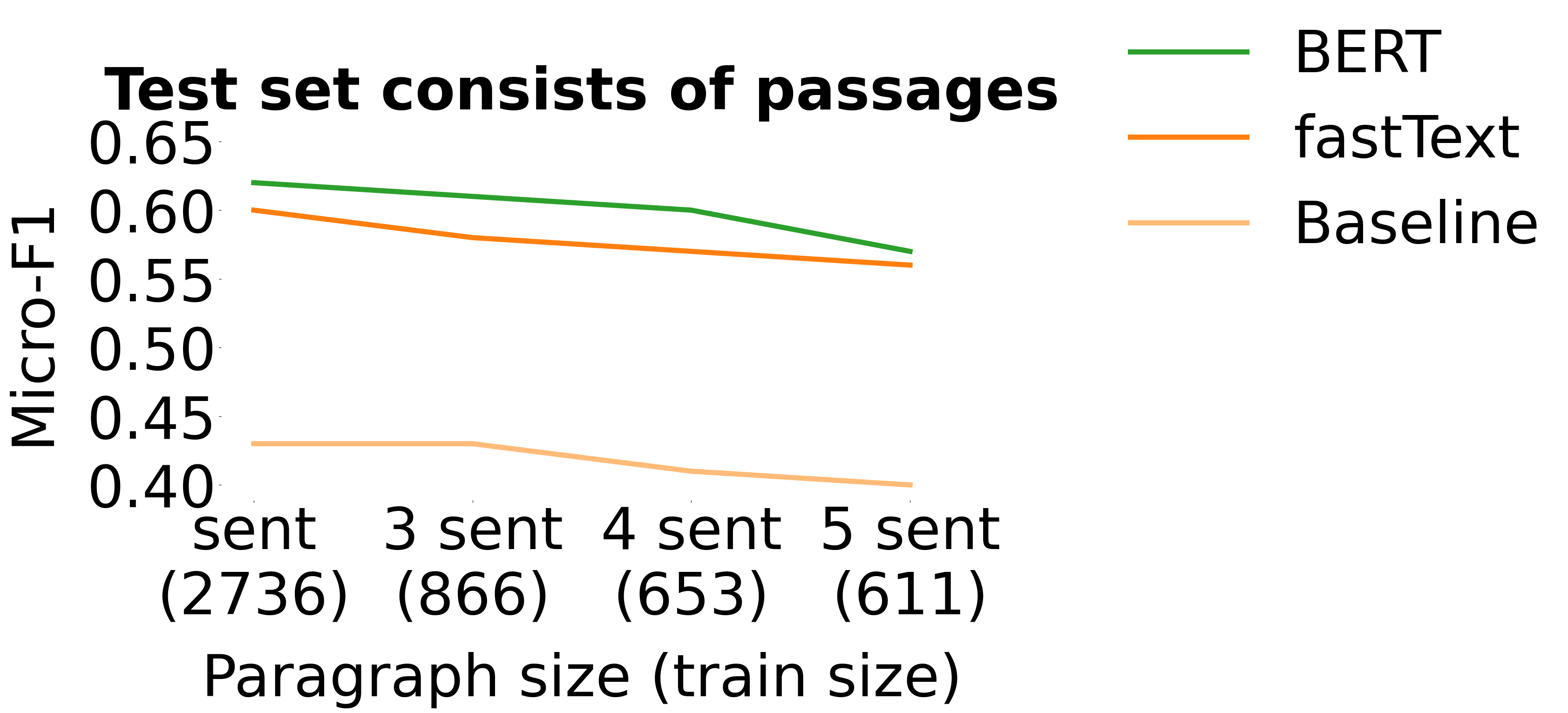}
           \caption{Micro-F1 measure per different paragraph size: test set consists of sentences(left); test set consists of paragraphs (right)}\label{fig4}
           
	      \end{center}
        \end{figure}

    In this section, we extend the analysis from Section \ref{ssec:machinevshuman} by looking at the effect of context versus the number of training instances provided for the classifier models. In this experiment, we gradually increase the length of the training instances in order to judge the importance of the training size versus the context (in terms of passage length). We evaluate the models using sentences and passages where each test passage consisted of three sentences (see Fig.~\ref{fig4}). The test sets for these experiments were extracted from the dev set while the training sentences and passages were extracted from the training set. Results showed that the performance of deep learning models is more influenced by the amount of the training instances rather than the length of the training passages. Further, models trained on sentence-level with a higher volume of training data give better results when tested on small paragraphs than classifiers trained on passages but with less training data available. This signifies the importance of higher volume of labelled data for reaching good classifiers performance.
    \vspace{-0.5cm}

\section{Conclusions and Future Work}\label{sec:conclusions}
\vspace{-0.1cm}
    Through this work, we explored the problem of predicting the main themes in safeguarding reports using supervised machine learning approaches. We analysed the performance of state-of-the-art classifiers, feature extraction and feature integration techniques which allowed us to identify classification methods suitable for domain-specific documents. Results showed that state-of-the-art deep learning model performance is highly dependent on the size of the training data in comparison to linear models as BERT's performance is worse than a simple Naive Bayes baseline and fastText for very small training datasets. Further, training word embeddings onto the specific domain, even when the size of the corpus is very small, lead to much higher results in comparison to pre-trained embeddings. This shows the importance of targeting pre-trained models to the specific corpus despite its size. The study comparing the expert validators' performance versus the automated models showed that the thematic analysis can be challenging even for subject-matter experts without prior knowledge of the thematic annotation framework. Further, humans need more knowledge about the context surrounding a sentence, compared to deep learning approaches. Experiments showed that BERT and fastText performance is more affected by the size of the training data rather than the amount of context given. On this respect, sentence-level classification provides more training data and fine-grained distinction between themes, which in turn allows for an easier expansion of the models and faster annotation.  

    In the future, we want to improve theme detection for the safeguarding documents by using generative language models for artificially augmenting the sparse data of the corpus. We will use the additional data as a training set in order to improve classifier performance. Further, we plan to look into developing and using knowledge graphs for improving classification. This will help refine the query functionality of the application and help improve the identification of similar documents and common trends in the safeguarding collection. 

\end{spacing}
\bibliographystyle{splncs04}
\bibliography{main}

\begin{thebibliography}{10}
\providecommand{\url}[1]{\texttt{#1}}
\providecommand{\urlprefix}{URL }
\providecommand{\doi}[1]{https://doi.org/#1}

\bibitem{ali2019text}
Ali, Z.: Text classification based on fuzzy radial basis function. Iraqi
  Journal for Computers and Informatics  \textbf{45}(1),  11--14 (2019)

\bibitem{bojanowski2017enriching}
Bojanowski, P., Grave, E., Joulin, A., Mikolov, T.: Enriching word vectors with
  subword information. Transactions of the Association for Computational
  Linguistics  \textbf{5},  135--146 (2017)

\bibitem{Devlin2019BERTPO}
Devlin, J., Chang, M.W., Lee, K., Toutanova, K.: Bert: Pre-training of deep
  bidirectional transformers for language understanding. ArXiv
  \textbf{abs/1810.04805}, ~16 (2019)

\bibitem{edwards2020go}
Edwards, A., Camacho-Collados, J., De~Ribaupierre, H., Preece, A.: Go simple
  and pre-train on domain-specific corpora: On the role of training data for
  text classification. In: Proceedings of the 28th International Conference on
  Computational Linguistics. pp. 5522--5529 (2020)

\bibitem{edwards2019knowledge}
Edwards, A., Preece, A., De~Ribaupierre, H.: Knowledge extraction from a small
  corpus of unstructured safeguarding reports. In: European Semantic Web
  Conference. pp. 38--42. Springer, Portoroź, Slovenia (2019)

\bibitem{fan2008liblinear}
Fan, R.E., Chang, K.W., Hsieh, C.J., Wang, X.R., Lin, C.J.: Liblinear: A
  library for large linear classification. Journal of machine learning research
   \textbf{9}(Aug),  1871--1874 (2008)

\bibitem{gururangan2020don}
Gururangan, S., Marasovi{\'c}, A., Swayamdipta, S., Lo, K., Beltagy, I.,
  Downey, D., Smith, N.A.: Don't stop pretraining: Adapt language models to
  domains and tasks. arXiv preprint arXiv:2004.10964  (2020)

\bibitem{joulin2017bag}
Joulin, A., Grave, E., Bojanowski, P., Mikolov, T.: Bag of tricks for efficient
  text classification. In: Proceedings of the 15th Conference of the European
  Chapter of the Association for Computational Linguistics: Volume 2, Short
  Papers. pp. 427--431. Association for Computational Linguistics, Valencia,
  Spain (April 2017)

\bibitem{mikolov2013distributed}
Mikolov, T., Sutskever, I., Chen, K., Corrado, G., Dean, J.: Distributed
  representations of words and phrases and their compositionality. Advances in
  Neural Information Processing Systems  \textbf{26},  3111--3119 (10 2013)

\bibitem{pedregosa2011scikit}
Pedregosa, F., Varoquaux, G., Gramfort, A., Michel, V., Thirion, B., Grisel,
  O., Blondel, M., Prettenhofer, P., Weiss, R., Dubourg, V., et~al.:
  Scikit-learn: Machine learning in python. Journal of machine learning
  research  \textbf{12}(Oct),  2825--2830 (2011)

\bibitem{radford2019language}
Radford, A., Wu, J., Child, R., Luan, D., Amodei, D., Sutskever, I.: Language
  models are unsupervised multitask learners. OpenAI Blog  \textbf{1}(8), ~9
  (2019)

\bibitem{robinson2019making}
Robinson, A.L., Rees, A., Dehaghani, R.: Making connections: a
  multi-disciplinary analysis of domestic homicide, mental health homicide and
  adult practice reviews. The Journal of Adult Protection  \textbf{21}(1),
  16--26 (2019)

\bibitem{sainz2021ask2transformers}
Sainz, O., Rigau, G.: Ask2transformers: Zero-shot domain labelling with
  pre-trained language models. arXiv preprint arXiv:2101.02661  (2021)

\bibitem{spasic2013flexiterm}
Spasi{\'c}, I., Greenwood, M., Preece, A., Francis, N., Elwyn, G.: Flexiterm: a
  flexible term recognition method. Journal of biomedical semantics
  \textbf{4}(1), ~27 (2013)

\bibitem{turker2019knowledge}
T{\"u}rker, R., Zhang, L., Koutraki, M., Sack, H.: Knowledge-based short text
  categorization using entity and category embedding. In: European Semantic Web
  Conference. pp. 346--362. Springer (2019)

\bibitem{wang2018glue}
Wang, A., Singh, A., Michael, J., Hill, F., Levy, O., Bowman, S.R.: Glue: A
  multi-task benchmark and analysis platform for natural language
  understanding. arXiv preprint arXiv:1804.07461  (2018)

\bibitem{Wolf2019HuggingFacesTS}
Wolf, T., Debut, L., Sanh, V., Chaumond, J., Delangue, C., Moi, A., Cistac, P.,
  Rault, T., Louf, R., Funtowicz, M., Brew, J.: Huggingface's transformers:
  State-of-the-art natural language processing. ArXiv  \textbf{abs/1910.03771}
  (2019)

\end{thebibliography}

\end{document}